# Scene Text Detection with Selected Anchors

Anna Zhu
Wuhan University of Technology
Wuhan, China
Annakkk@live.com

Hang Du
Wuhan University of Technology
Wuhan, China
duhangzz@outlook.com

Shengwu Xiong
Wuhan University of Technology
Wuhan, China
xiongsw@whut.edu.cn

*Abstract*—**Object proposal technique with dense anchoring scheme for scene text detection were applied frequently to achieve high recall. It results in the significant improvement in accuracy but waste of computational searching, regression and classification. In this paper, we propose an anchor selection-based region proposal network (AS-RPN) using effective selected anchors instead of dense anchors to extract text proposals. The center, scales, aspect ratios and orientations of anchors are learnable instead of fixing, which leads to high recall and greatly reduced numbers of anchors. By replacing the anchor-based RPN in Faster RCNN, the AS-RPN-based Faster RCNN can achieve comparable performance with previous state-of-the-art text detecting approaches on standard benchmarks, including COCO-Text, ICDAR2013, ICDAR2015 and MSRA-TD500 when using single-scale and single model (ResNet50) testing only.**

## I. Introduction

Text information in natural scenes provides important supplementary clue for content-based image understanding and analysis. They can identify and indicate objects and location when correctly read. Recently, scene text detection and recognition have gained increasing attention in computer vision community due to its practical and potential applications, such as license plate recognition, automatic translation, robot navigation, virtual reality, etc. Text detection is the initial and promising step in general processing pipeline for its accuracy has extremely influence on the sequential text recognition. However, the challenges including the various text appearance and layout, non-uniform illumination, blurring, low resolution, noise, and background clutters effect the text detection accuracy drastically[1].

Over the last decade, the traditional text detection methods have promoted the development through two pipelines [2][3]. One is the connected component analysis (CCA)-based procedure and the other is sliding window-based procedure However, scene text detection in natural image is a high-level visual task, which is difficult to be solved completely by a set of low-level operations or manually designed features.

Deep neural networks (DNN), as in most common vision problems, have achieved remarkable performance in text detection as well [4][5]. It is capable of learning meaningful high-level features and semantic representations for visual recognition through a hierarchical architecture with multiple layers of feature convolutions. Hence, classifying and regressing the object or text proposals from a set of dense predefined region anchors becomes one essential way for detectors. For most text detection methods, such as SSD-based and RPN-based, the anchors are defined in a uniform scheme, where every location in a feature map is associated with fixed anchors with predefined scales, aspect ratios and orientations. However, most of them correspond to false candidates since text occupies limited regions in natural scene images. Meanwhile, large number of anchors can lead to significant computational cost during classification and regression stage. Additionally, it is difficult to design anchors adapting to various text instances.

In this paper, we use an anchor selection-based region proposal network (AS-RPN) to replace the original sliding anchor-based Region Proposal Network (RPN) in the Faster RCNN framework [6] for text detection. The AS-RPN extract anchors on three branches: the anchor location prediction branch, the anchor orientation estimation branch and the anchor shape prediction branch. The orientation estimation of the anchors ensures the AS-RPN can detect arbitrary orientation text. Anchor shape contains the width and height of text regions. It is predicted in each location of the feature map, and only one pair value will be assigned to each location. In this way, the selected anchors on the feature maps become sparse and will be mostly occupied on text regions.

The proposed method uses effective selected rotate anchors for accurate classification and regression. Scales, aspect ratios and orientations of the anchors are learnable now instead of fixed. Therefore, the anchor numbers are reduced drastically comparing to dense anchors. On the other hand, the anchors predicted by in this way are very effective for text detection which leads to high recall.

The main contributions of this work are presented as follows:

The text proposal extraction network AS-RPN implements a novel selected anchor scheme with the ability to predict non-uniform and arbitrary shaped anchors other than dense and predefined ones. The ablation experiment on COCO-Text [7] dataset illustrates this network can predict the text proposals with significant reduced anchors and can extract high-quality text proposals effectively.

The AS-RPN can detect text with arbitrary orientation. The experimental results on several benchmark datasets as COCO-Text, ICDAR2013 [8], ICDAR2015 [9] and MSRA-TD500 [10] demonstrate our proposed method can retain the high recall and also improve the precision extremely. Compared with current state-of-the-art text detection methods, it also shows its superiority.

## II. RELATED WORK

Scene text detection approaches are conventionally and typically classified to two categories: connected component analysis (CCA)-based and regional based approaches. Recently, even the scene text detection steps to a new era[11]after applying deep learning, the DNN-base algorithms can still be grouped to these two categories.

The connected component analysis-based methods detect text from pixel level to character level, then to text line level sequentially. Conventionally, the CCA-based approaches group pixels to candidate characters by low level features, and then grouped candidate character components are further verified.

Text lines are formed by grouping verified characters with heuristic rules at last. The Stroke Width Transform (SWT) [12] and Maximally Stable Extremal Regions (MSERs) [13] are two representative ones. Since CNN has the good ability to represent text features, it is widely incorporated with the CCA-based methods for text/non-text classification [14][15]. Recently, scene text detection trends to design the end-to-end deep learning models. The CCA-based text detection methods initially predict the text probability of each pixel by Fully Convolutional Network(FCN)-based models[16][17]. Those methods cast text detection as segmentation problem. High-level features are extracted directly from the whole images for text regions detection. Algorithms directly runs on full images and few post processing steps are required. Compared with the conventional CCA-based method, they are more robust and faster.

Regional-based methods usually adopt sliding window strategy casting the text detection as a classical object detection task. By scanning the images with multi-scale windows, discrete sub-image space regions are captured, and then classified through texture classifiers or CNN. Wang et al. [18]use a random ferns classifier trained on HOG features in a sliding window scenario to find characters in an image. Wang et al. [19] normalized all the extracted regional patches to input into a CNN classifier for character patch classification. Jaderberg et al.[20] combined of an object-agnostic region proposal method and a sliding window detector to generate word candidates. Extracting feature for each region independently was identified as the bottleneck in these exhaustive regions searching manner.

The object proposals techniques which share convolutions across proposals[21][22] carried on with the deep neural networks emerge as an alternative to the conventional regional-based text detection approaches. Text location regression is performed on referenced sub-regions of specified layers of DNN, which correspond to initial regions to be predict. Zhong et al. [23] designed an inception region proposal network to achieve only hundred level candidate text proposals from a set of text characteristic prior bounding boxes. Those candidate text regions are further classified and regressed for accurate localization by a text detection network that embeds ambiguous text category information and multilevel region-of-interest pooling.

Gupta et al. [24] proposed a fully-convolutional regression network, which draws on the image-grid based bounding box regression network YOLO [21], to perform text detection at all locations and multiple scales in the image. Liao et al. [25] designed a FCN-based TextBoxes model that directly outputs the coordinates of word bounding boxes at multiple network layers by jointly predicting text presence and coordinate offsets to a set of default regions on multiple layers. Furthermore, various deep neural networks are adaptively modified to detect arbitrary-oriented scene text by designing multi-oriented anchors [26], linking SSD-predicted text segments to complete text [27], or regressing corner points of text regions [28]. These kinds of methods take advantage of the development in general object detection. Instead of cropping the image to a number of sub-regions and evaluation the CNN thousands of times per image, these latest works regress the text region coordinates on sparse proposals and detect text on a single forward pass. Additionally, these modules can be trained end-to-end for accurate text detection.

## III. THE PROPOSED METHED

Most current regional-based text detection methods adopt anchor scheme which rely on designing a set of uniform arrangement of size fixed anchors. Many of the anchors are placed in regions where the text of interest is unlikely to exist and different sizes of anchors at the close positions return to the same text regions with high overlapping ratios. Therefore, we turn to use selected anchors [29] to adapt the text region of interest.

### A. Anchor-selected Region Proposal Network

We adopt the AS-RPN based Faster RCNN for text detection. The architecture is shown in Fig. 1. It is composed of three components, namely the Feature Pyramid Network (FPN) [30], anchor selection network and text prediction network. The FPN was use for feature extracting at different levels. Since text size varies greatly in natural scene images, we develop a multi-level anchor generation scheme, which collect anchors at multiple feature maps for detection. This feature extraction network follows the FPN architecture [30] which uses the Resnet101 and the FPN as backbone and adopts the features in the layer of $\{P_2, P_3, P_4, P_5\}$.

The inputs of the network are the whole images and directly outputs regressed text bounding boxes and text confidence from the convolutional features, referring to a set of predicted anchors with arbitrary locations, orientations and shapes.

Intuitively, the feature for a large anchor should encode the content over a large region, and those for small anchors should have smaller scopes accordingly. Therefore, we implement the deformable convolution [31] to transform the features at location based on the anchor shapes. On the top of the transformed features, the text classification and bounding-box regression are performed for extracting text proposals. After that, the Rotation Region-of-Interest (ROI) pooling [26] and accurate text prediction in Faster RCNN are adopted for identifying text regions.

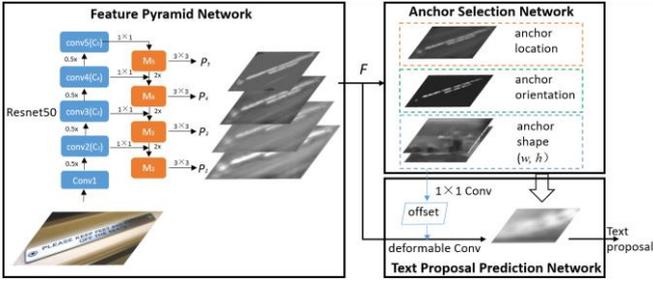

Fig.1. The architecture of the AS-RPN based text detection. The feature pyramid network built on Resnet101 and the FPN is used as the backbone network to extract features. The anchor selection network estimates the anchor location, orientation and shape. To get the textness scores and regressed bounding boxes, text prediction network uses deformable convolution based on the anchor shapes for feature normalization.

*B. Anchor Prediction*

The anchor selection network goes through three streams for predicting the anchor location, the orientation and the shape of width and height at each location. The anchor location prediction branch outputs a probability map of the same size as the input feature map, which indicates the probability of the text center existing at that location. It applies a $1 \times 1$ convolution to the input feature map to obtain a map of text scores, which are then converted to probability values via an element-wise sigmoid function. A threshold $T_a$ is applied on the resultant probability map to determine the active anchor locations. It controls the sparsity of anchor distribution. Experimentally, we set it to 0.05 for it can filter out more than 90% of the regions while still maintaining a high recall.

The anchor orientation branch estimates the text orientation $\theta$ in each location. It outputs a soft map in which each foreground pixel is assigned with to a value within the range of [0,1]. It corresponds to orientation $\theta$ in the range of $[-\pi/2, \pi/2]$. The real orientation of text is mapped to [0,1] by shifting and normalization. The orientation values beyond this range is converted to it.

For the anchor shape prediction branch, it is performed on feature map of each level and predicts the best shape $(w, h)$ which has the highest IoU with the ground truth bounding box at each location. The shape is estimated by Eq. (1). Instead of estimating the w and h, the shape prediction branch will output $dw$ and $dh$ and then be transformed to $(w, h)$ as above, where $s$ is the stride and $k$ is an empirical scale factor ($k = 5$ in our experiments). This process outputs space from approximate [0, 500] to [-1, 1], leading to an easier and stable learning target. The shape prediction branch also applies a $1 \times 1$ convolution to the input feature map and outputs two-channel map indicating the offset of $dw$ and $dh$ respectively. The anchor shape prediction differs significantly from the conventional anchoring schemes in that every location is associated with just one anchor of the dynamically predicted shape instead of a set of anchors of predefined shapes.

$$w = k \times s \times exp(dw), h = k \times s \times exp(dh) \quad (1)$$

After we get the anchor location, orientation and shape, text proposal prediction of the text-ness score and bounding box can be implemented. Since the predicted anchors have various shapes, to process the feature normalization becomes the key point for prediction. Instead of using the fully convolutional classifier uniformly over the feature map, we applying deformable convolution to transform the features at individual location based on the anchor shapes. In detail, we first predict an offset field from the output of anchor shape prediction branch, and then apply the $3 \times 3$ deformable convolution to the original feature map with the offsets to obtain the transformed feature maps. The text proposal classification and bounding-box regression is performed based on the transformed feature map and the selected anchors.

*C. Label generation*

For each training sample with the input image $I$ and the corresponding ground truth, we generate location targets, angle targets and shape targets for AS-RPN and Faster RCNN. Ideally, the ground truth files should be constituted by the 5-tuple in the form of $R\{x_g, y_g, w_g, h_g, \theta_g\}$, where $(x_g, y_g)$ is the center coordinate of the instance, $w_g$ is the width, $h_g$ is the height and $\theta_g$ presents the rotate angle of the bounding box. Different ground truth can be transferred in the following way.

**Anchor location targets.** To train the anchor location branch, the binary label map with 1 represents a valid anchor and 0 otherwise is required. We employ ground-truth bounding boxes for guiding the binary label map generation. Pixels in a shrink region with scale pair $(\sigma_1, \sigma_2)$ (namely the region $\sigma_1 \times w_b, \sigma_2 \times h_b$) within the ground truth bounding box $(w_b, h_b)$ are defined as positive anchor location. The other pixels within the ground-truth bounding box are marked as "ignore". The pixels laying outside of the ground-truth bounding boxes are negative samples.

**Anchor orientation targets.** To generate the best orientation target for the multi- orientated text detection challenges, data augment and angle define are required. In most datasets, the ground truth bounding boxes are defined by four vertices $(x_1, y_1, ..., x_4, y_4)$. It can be converted to the rotate angle of the bounding boxes by the following algorithm 1:

| Algorithm1 angle label generation | |
|---|---|
| 1: | Input: original gt $O(x_1, y_1, ..., x_4, y_4)$, $A(x_1, y_1)$, $B(x_2, y_2), C(x_3, y_3), D(x_4, y_4)$ as shown in Fig.2 |
| 2: | Output: output gt $F(x, y, w, h, \theta)$ |
| 3: | for each line in O, do |
| 4: | $(x, y) = (\frac{x_1+\cdots+x_4}{4}, \frac{y_1+\cdots+y_4}{4})$ |
| 5: | $w = \max(AB, AD), h = \min(AB, AD)$ |
| 6: | Calculate the middle point{E,F,G,H} of AB, BC, CD and DE |
| 7: | $\theta_1 = \arctan k_{EG}, \theta_2 = \arctan k_{HF}$ |
| 8: | if len(EG) > len(HF) |
| 9: | $F = (x, y, w, h, \theta_1)$ |
| 10: | else $F = (x, y, w, h, \theta_2)$ |
| 11: | end if |
| 12: | end for |

For the horizontal text regions in ICDAR 2013 and COCO-Text, we rotate the image by an random angle $\theta_0 \in [-\pi/2, \pi/2]$

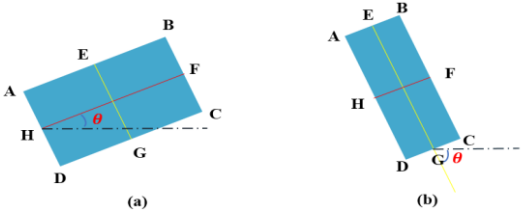

Fig. 2. Examples of the orientation of the bounding boxes: (a) wider box (b) higher box. The orientation of (b) is converted to $[-\pi/2, \pi/2]$.

around its center, and the center of the anchor can be calculated as in Eq. (2).

$$\begin{bmatrix} x' \\ y' \\ 1 \end{bmatrix} = T(\tfrac{I_w}{2}, \tfrac{I_h}{2}) R_{\theta_0} T(-\tfrac{I_w}{2}, -\tfrac{I_h}{2}) \begin{bmatrix} x \\ y \\ 1 \end{bmatrix}, \quad (2)$$

where $I_w$ and $I_h$ is the original text region, $T$ and $R$ are the translation and rotation matrix as depicted in Eq. (3) and (4), respectively.

$$T(a, b) = \begin{bmatrix} 1 & 0 & a \\ 0 & 1 & b \\ 0 & 0 & 1 \end{bmatrix} \quad (3)$$

$$R(\theta) = \begin{bmatrix} \cos\theta & \sin\theta & 0 \\ -\sin\theta & \cos\theta & b \\ 0 & 0 & 1 \end{bmatrix} \quad (4)$$

We donate $\theta_g \in (-\tfrac{\pi}{2}, \tfrac{\pi}{2})$ as the angle target in the region whose center is $(x_g, y_g)$ and the size of $w_g \times h_g$, The orientation values beyond this range is converted to it. Then the orientation targets are mapped to (0, 1) by shifting and normalization as shown in Eq. (5).

$$\theta_t = \frac{\theta_g}{\pi} + \frac{1}{2} \quad (5)$$

For some pixels may covered by bounding boxes with different orientations, the angle target of these pixels are set to the averaged value of the ground truth orientations of the corresponding locations.

**Anchor shape targets.** To determine the shape target for each anchor, we need to match the anchor to a ground truth bounding box and the compute the optimal shape $w$ and $h$ which can maximize the IoU between the anchor and the matched ground truth bounding box as max IoU (anchor$(w, h)$, $gt$). $(w, h)$ is the shaper of a certain anchor and $gt$ is the ground-truth bounding boxes. It is hard to be implemented efficiently for jointly optimization. Alternatively, we sample some common values of w and h to simulate the enumeration of the two variables. Then we calculate the IoU of these sampled anchors with $gt$ to find the maximum approximation.

*D. Network Training*

The proposed framework is optimized by using a multi-task loss which contains the text confidence loss $L_{conf}$, text localization loss $L_{reg}$ and an additional anchor loss. The anchor loss can be further divided into localization loss $L_{loc}$, orientation loss $L_{angle}$ and shape prediction loss $L_{shape}$. They are jointly used to optimize the detection model as defined in Eq. (6).

$$L = L_{conf} + L_{reg} + \alpha L_{loc} + \beta L_{angle} + \lambda L_{shape} \quad (6)$$

We use cross entropy loss for the classification of text or non-text, and apply the smooth-$L_1$ loss for regressing the offset of the width, height and centers coordinates of each bounding box referring to the selected anchors.

**Anchor location loss.** As shown in Fig 1, the output of the location prediction branch is a probability mask $p(i, j|F_I)$ of the same size as the input feature map $F_I$, where $(i, j)$ corresponds to the location with coordinate $\left(\left(i + \tfrac{1}{2}\right)s, \left(j + \tfrac{1}{2}\right)s\right)$ on $I$.

Since the center of text instance usually accounts for a small area of the whole feature map, we use Focal Loss to train the location prediction branch as shown in Eq.(7).

$$L_{loc} = \begin{cases} -\alpha(1 - y')^\gamma \log y', & y = 1 \\ -(1 - \alpha) y'^\gamma \log(1 - y'), & y = 0 \end{cases} \quad (7)$$

where $y'$ is the output of the anchor location branch with a sigmoid function.

**Anchor orientation loss.** Similar to the location prediction branch, the output of the orientation prediction branch is a probability map with the size of $b \times c \times w \times h$. Where $b$ is the batch size, $c$ is the channel, and $w$ and $h$ are the width and height of the feature map, respectively. We adopt a modified cosine loss as defined in Eq. (8) to optimize the orientation prediction branch. $\hat{\theta}$ is the prediction of the orientation branch and $\theta_g$ is the angle target.

$$L_{angle} = 1 - \cos(\hat{\theta} - \theta_g) \quad (8)$$

**Anchor shape loss.** Different from the conventional anchor-based methods, the shape prediction branch does not change the anchor positions which will never cause the misalignment between anchors and anchor features. The goal is to predict the values of $w$ and $h$, while the output is $d_w$ and $d_h$ and then can be mapped to $w, h$ with Eq. (1).

We adopt a bounded IoU loss to optimize the shape prediction branch. The loss is defined as follows:

$$L_{shape} = L_1\left(1 - min\left(\frac{w}{w_g}, \frac{w_g}{w}\right)\right) + L_1\left(1 - min\left(\frac{h}{h_g}, \frac{hg}{h}\right)\right) \quad (9)$$

Where $(w, h)$ denote the predicted anchor shape, $(w_g, h_g)$ is the shape of the corresponding ground-truth bounding box. $L_1$ is the smooth $L_1$ loss

IV. DATASET

We evaluate our approach on four benchmark datasets: COCO-Text, ICDAR2013, ICDAR2015 and MSTD500.

**COCO-Text.** The dataset contains 63,686 images with 173,589 labeled text regions. For each text region, it provides the location in terms of bounding boxes, classifications in terms of legibility, category (e.g. machine printed or hand written). For our text detection usage, we select around 10,000 machine printed text images with full bounding box annotation for validation. It is mainly used for training and proposal quality evaluation.

**ICDAR2013.** This dataset contains 229 training images and 233 test images which are well captured in high resolution. Texts in the images of this dataset are mostly horizontal and focused. In all the training images, 848 text regions are extended cropped for training the text relocation system.

**ICDAR 2015.** It is released for the text localization of incidental scene text challenge, which has 1500 images in total. 1000 of them are used for training and the remaining are for testing the text are annotated with irregular quadrilateral bounding box vertices with orientation information. It is used for multi-oriented text detection.

**MSRA-TD500.** It is a dataset comprises of 300 training images and 200 test images. Text regions are of arbitrary orientations. It contains text in both English and Chinese. The text regions are annotated in RBOX format.

## V. EXPERIMENTAL RESULTS

### A. Implementation Details.

The backbone network we used is ResNet-101 with FPN and the images were resized to $1333 \times 800$, without changing the aspect ratio. The model is optimized by stochastic gradient descent (SGD) over 4 1080Ti GPUs. The AS-RPN adopts 16 images per mini batch with a weight decay of 0.0001 and momentum of 0.9. The weights of the network are updated by using a learning rate of 0.002 for the first 10 epochs, and decrease by 0.1 at epoch 10 and 14 (16 epochs in total).

We set $\sigma_1 = 0.4$ and $\sigma_2 = 0.5$ for anchor location prediction. In the multi-task loss function (9), we simply set $\alpha = \beta = 1$ and $\lambda = 0.1$ to balance the location, orientation and shape prediction branches. Four sizes of scale {8, 16, 32, 64} are sampled with three aspect ratios {1,2,4} at each position on each pyramid level in {$P_2, P_3, P_4, P_5$}. Since words tend to have large aspect ratios, we sample three additional "long" anchors with aspect ratios {3, 5, 7} in {$P_2, P_3$}. After the AS-RPN is well trained, we use Rotation ROI pooling to align the features in text proposals and integrate it with converged Faster RCNN by removing the RPN and fine tune the whole text detector with training data from all datasets for 5000 iterations.

Only a polygon non-maximum suppression (NMS) [26] was used for the post processing to obtain the bounding boxes as the final result. We rescore all text instances and find the minimum bounding rectangle for each text instance. The polygon NMS is utilized to suppress redundant boxes. We use the rotation of the images with a random angle in   for the data augmentation and re-calculate the coordinates of the annotated bounding boxes with function (2)-(4). Compared to the performance without augmentation, the efficiency can improve by 18.5%, 21%, 19.7% in precision, recall and F-measure.

### B. The effect of each branch of AS-RPN

We investigated the importance of angle prediction branch and the effect on the decrease of anchor number.

**Anchor location.** The location branch leads to more efficient inference because of its usefulness of obtaining sparse anchors. The threshold $T_a$ of this branch controls the distribution of anchors as shown in Tab. I. A smaller $T_a$ can greatly reduce the number of anchors, but only a minor influence on the instance recall rate. With this design, we can reduce anchors by more than 90%, and lead to lighter NMS in the post processing. We set $T_a$ to 0.05 in our method for tradeoff between the computational cost effecting by anchor numbers and the evaluation performance.

TABLE I. THE EFFECTS OF DIFFERENT $T_a$ ON ANCHOR NUMBERS ON MSRA-TD500 WITH IoU THRESHOLD = 0.5

| Ta | Anchors/image | P (%) | R (%) | F-measure (%) |
|---|---|---|---|---|
| 0 | 108864(100.0%) | 84.73 | 80.42 | 82.52 |
| 0.01 | 32310(29.7%) | 84.73 | 80.35 | 82.51 |
| 0.05 | 6858(6.3%) | 84.67 | 80.37 | 82.49 |
| 0.1 | 3266(3.0%) | 82.35 | 78.20 | 80.22 |

**Anchor shape.** General dense anchor methods pre-define $k$ anchors ($k$ =54 in RRPN) of different scales and aspect ratios, while our method predicts only one anchor at each cell of the feature map, which reduces the anchor number to $1/k$. However, the competitive F-measure of 82.49% (RRPN is 74%) proves anchors designed by our method can provide more accuracy candidates on text detection.

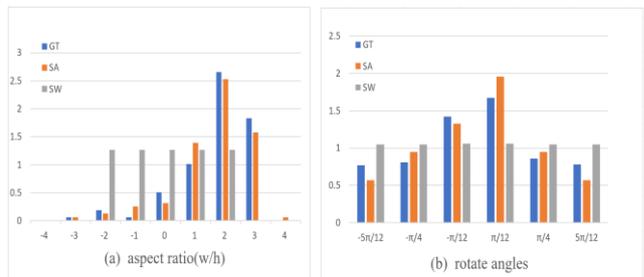

Fig. 3. (a) Anchor aspect ratio and (b) rotate angles distribituions of different anchoring schemes. The x-axis of (a) is reduced by $log_2^{(\cdot)}$. GT, SA, SW stands for ground truth, selected anchor and sliding window.

**Anchor angle.** The angle prediction branch improves the recall by a large scale for multi-orientation text detection. By adding this branch to the pipeline, the results of precision, recall and F-measure improve from 53.36%, 56.20%, 54.67 to 84.67%, 80.37%, 82.49%. Results shown in Fig. 3. also prove that the predicted anchors cover a wider range of scales, aspect ratios and orientations, which can provide a more similar distribution to Ground truth.

### C. Region Proposal Quality Evaluation

We then observe the performance of AS-RPN for text proposal generation. Since this network aims to extract high-quality text proposals, we use the text recall (TR) to evaluate its performance with limited number of proposals. TR for 50, 100 and 300 proposals per image are denoted as TR50, TR100 and TR300. They are computed at a single IoU threshold of 0.5, 0.75 and averaged at multiple IoU thresholds between 0.50 and 0.95 with an interval of 0.05, respectively. A series of ablation experiments are conducted to compare AS-RPN to RPN [6], FPN-RPN [30] and anchor free (AF)-RPN [32] in region proposal generation task on the COCO-Text dataset.

Specifically, anchors with 4 scales {8, 16, 32, 64} and 5 aspect ratios {0.2, 0.5, 1.0, 2.0, 5.0} at each sliding position on $C_4$ are used for RPN. AF-RPN is based on Faster RCNN framework incorporating the anchor-free idea of DenseBox. It directly predicts the offsets from sliding points to the bounding box vertices of the concerned text instance. Instead of design the shape arbitrary anchors, the sliding points can be considered as alternative anchors. In the training stage, an anchor is assigned a positive label if it has an IoU overlap higher than 0.5 with any ground-truth boxes or the highest IoU for a given ground truth box and a negative label if it has an IoU less than 0.1 for all ground truth boxes. The results are listed in Tab. II.

TABLE II. REGION PROPOSAL QUALITY EVALUATION ON COCO-TEXT VALIDATION SET (%)

| Measure \ Method | | RPN | FPN-RPN | AF-RPN | AS-RPN |
|---|---|---|---|---|---|
| IoU_0.5 | $TR_{50}$ | 67.2 | 67.5 | 73.3 | **74.5** |
| | $TR_{100}$ | 76.9 | 77.2 | 81.8 | **82.9** |
| | $TR_{300}$ | 86.6 | 87.4 | **89.3** | 88.6 |
| IoU_0.75 | $TR_{50}$ | 22.8 | 28.8 | 35.0 | **36.2** |
| | $TR_{100}$ | 27.9 | 36.0 | 41.3 | **44.6** |
| | $TR_{300}$ | 33.8 | 47.2 | 48.2 | **48.8** |
| IoU_Avg | $TR_{50}$ | 30.6 | 33.5 | 38.2 | **38.8** |
| | $TR_{100}$ | 35.9 | 39.8 | 43.6 | **44.9** |
| | $TR_{300}$ | 41.7 | 48.0 | 49.2 | **50.0** |

It can be seen that our proposed AS-RPN outperforms the other three methods, especially the methods of RPN and FPN-RPN. This result demonstrates the effectiveness of our proposed AS-RPN. The recall improvements are more significant if we keep 100 proposals. Some examples of text proposals generated upon sliding window anchoring FPN-RPN and our selected anchoring are displayed in Fig. 4. It is observed that the high-quality text proposals concentrate more on text by our AS-RPN, and the numbers reduced more comparing to designed sliding window-based anchor scheme.

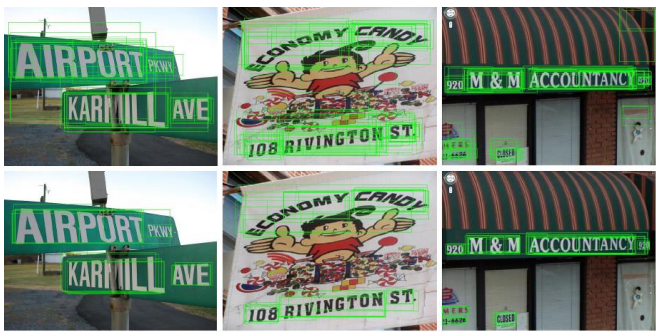

Fig. 4. Examples of FPN-RPN text proposals (top row) and AS-RPN text proposals(bottom row).

D. Comparing with state-of-the-art approaches

we compare our method with other relevant state-of-the-art scene text detection methods for better understanding the superiority of our approach.

**Horizontal text.** Firstly, we evaluate our method on ICDAR2013 dataset to verify its effectiveness on horizontal scene text detection. The results of our model are listed and compared with current state-of-the-art method in Tab. III. Comparing with the anchor based detection methods and some segmentation based methods, our model outperforms most of them with the precision of 90.19%, recall of 91.16% and F-measure of 90.62%.

TABLE III. DETECTION RESULTS COMPARE WITH RELAVENT APPROACHES ON ICDAR 2013

| Approach | P(%) | R(%) | F(%) |
|---|---|---|---|
| CPTN[35] | 74.22 | 51.56 | 60.85 |
| Seg Link[27] | 74.74 | 76.50 | 75.61 |
| SSTD[41] | 80.23 | 73.86 | 76.91 |
| RRPN[26] | 82.02 | 73.00 | 77.05 |
| EAST*[38] | 84.36 | 81.27 | 82.79 |
| R2CNN[42] | 85.62 | 79.68 | 82.54 |
| Text boxes++[25] | 87.80 | 78.50 | 82.90 |
| **Ours** | **83.34** | **79.99** | **81.63** |

**Oriented text.** To verify the superiority of our method in oriented quadrangle text, we conduct experiments on ICDAR 2015 and MSRA-TD500 datasets. In ICDAR2015, the results submitted online achieves the recall of 83.34%, precision of 79.99% and F-measure of 81.63% as shown in Tab. IV.

TABLE IV. DETECTION RESULTS COMPARE WITH RELAVENT APPROACHES ON ICDAR 2015

| Approach | P(%) | R(%) | F(%) |
|---|---|---|---|
| Faster-RCNN[6] | 75.00 | 71.00 | 73.00 |
| Text boxes[40] | 87.73 | 82.59 | 85.08 |
| Yao et al[33] | 88.91 | 80.20 | 84.32 |
| SSTD[34] | 89.00 | 86.00 | 88.00 |
| RRPN[26] | 90.12 | 72.00 | 80.45 |
| CTPN[35] | 92.02 | 83.88 | 88.45 |
| Mask TextSpotter[36] | 88.27 | 94.01 | 90.82 |
| **Ours** | **90.19** | **91.16** | **90.62** |

A high recall of detection is obtained. For MSRA-TD500 dataset, we evaluate the results using the same evaluation method with RRPN. As shown in Tab. V, we can see that the precision of 84.67%, recall of 80.36% and F-measure of 82.46% are achieved by our method and it outperforms other competitors significantly.

TABLE V. DETECTION RESULTS COMPARE WITH RELAVENT APPROACHES ON MSRA-TD500

| Approach | P(%) | R(%) | F(%) |
|---|---|---|---|
| Baseline | 57.40 | 54.50 | 55.90 |
| He et al[37] | 76.40 | 61.42 | 68.76 |
| EAST*[38] | 81.23 | 63.27 | 75.54 |
| RRPN[26] | 68.00 | 82.00 | 74.00 |
| TextSnake[39] | 83.20 | 73.90 | 78.30 |
| Pixel Link[17] | 83.00 | 73.20 | 77.82 |
| Lyu et al[28] | 87.60 | 76.20 | 81.50 |
| **Ours** | **84.67** | **80.37** | **82.49** |

The competitive results on both ICDAR 2015 and MSRA-TD500 validated the effectiveness and robustness of our method in the mainstream oriented text detection tasks. As shown in Fig. 5, our text detector can detect scene text regions under various challenging conditions, such as low-resolution, complex background, large aspect ratios as well as varying orientation.

There are some limitations in our approach, for example, the method is sensitive to the predicted rotation angles, a $\pi/15$ error of a bounding box whose aspect ratio is 5:1 will cause the IoU decrease to 0.4.

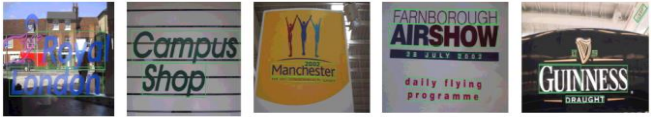

(a) Detection results in ICDAR2013

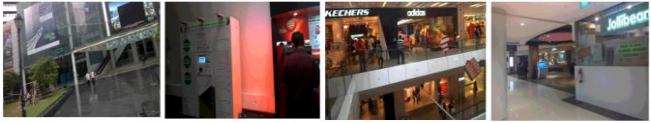

(b) Detection results in ICDAR2015

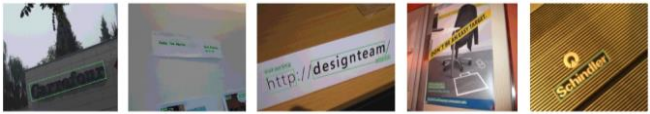

(d) Detection results in MSRA-TD500

Fig.5 Some detection results of our method on different benchmarks.

## CONCLUSION

In this paper, we introduced an accurate scene text detection approach which was inherited from a Faster RCNN by replacing RPN with AS-RPN. The AS-RPN generated high-quality text proposals through anchor location prediction, anchor orientation estimation and anchor shape prediction branches. Experimental results on COCO-Text, ICDAR2013, ICDAR2015 and MSRA-TD500 datasets demonstrated the superiority of our proposed method. Additionally, the AS-RPN module can be concatenated with other text detection methods to boost their text location accuracy and efficiency.

ACKNOWLEDGMENT

This work was partly supported by the National Natural Science Foundation of China under Grant 61703316 and Independent innovation program for college students No.195210012.

REFERENCES

[1] Y. Zhu, C. Yao, X. Bai, Scene text detection and recognition: Recent advances and future trends, Frontiers of Computer Science 10 (1) (2016) 19–36.

[2] Q. Ye, D. Doermann, Text detection and recognition in imagery: A survey, IEEE Trans. PAMI 37 (7) (2014) 1480–1500.

[3] X.-C. Yin, Z.-Y. Zuo, S. Tian, C.-L. Liu, Text detection, tracking and recognition in video: A comprehensive survey, IEEE Transactions on Image Processing 25 (6) (2016) 2752–2773.

[4] T. He, Z. Tian, W. Huang, C. Shen, Y. Qiao, C. Sun, An end-to-end textspotter with explicit alignment and attention, in: Proceedings of International Conference on Computer Vision and Pattern Recognition, 2018.

[5] X. Liu, L. Ding, Y. Shi, D. Chen, Q. Yu, J. Yan, Fots: Fast oriented text spotting with a unified network, in: Proceedings of International Conference on Computer Vision and Pattern Recognition, 2018.

[6] S. Ren, K. He, R. Girshick, J. Sun, Faster r-cnn: towards real-time object detection with region proposal networks, IEEE Transactions on Pattern Analysis and Machine Intelligence 39 (6) (2017) 1137–1149.

[7] Andreas Veit, et al. COCO-Text: Dataset and Benchmark for Text Detection and Recognition in Natural Im-ages. arXiv preprint arXiv:1601.07140, 2016.

[8] Dimosthenis Karatzas, et al. ICDAR 2013 robust reading competition. In ICDAR, pp. 1484-1493, 2013.

[9] Dimosthenis Karatzas, et al. ICDAR 2015 robust reading competition. In ICDAR, pp. 1156-1160, 2015.

[10] C. Yao, X. Bai, W. Liu, Y. Ma, and Z. Tu. Detecting texts of arbitrary orientations in natural images. In CVPR, pages 1083–1090, 2012.

[11] S. Long, X. He, C. Yao, Scene text detection and recognition: The deep learning era, arXiv preprint arXiv:1811.04256.

[12] B. Epshtein, E. Ofek, Y. Wexler, Detecting text in natural scenes with stroke width transform, in: Proceedings of International Conference on Computer Vision and Pattern Recognition, 2010, pp. 2963–2970.

[13] J. Matas, O. Chum, M. Urban, T. Pajdla, Robust wide-baseline stereo from maximally stable extremal regions, Image and vision computing 22 (10) (2004) 761–767.

[14] W. Huang, Y. Qiao, X. Tang, Robust scene text detection with convolution neural network induced mser trees, in: Proceedings of European Conference on Computer Vision, 2014, pp. 497–511.

[15] A. Zhu, R. Gao, S. Uchida, Could scene context be beneficial for scene text detection?, Pattern Recognition 58 (2016) 204–215.

[16] M. Liao, Z. Zhen, B. Shi, G. S. Xia, B. Xiang, Rotation-sensitive regression for oriented scene text detection, in: Proceedings of International Conference on Computer Vision and Pattern Recognition, 2018.

[17] D. Dan, H. Liu, X. Li, C. Deng, Pixellink: Detecting scene text via instance segmentation, in: Proceedings of International Conference on AAAI, 2018.

[18] K. Wang, B. Babenko, S. Belongie, End-to-end scene text recognition, in: Proceedings of International Conference on Computer Vision, 2011, pp. 1457–1464.

[19] T. Wang, D. J. Wu, A. Coates, A. Y. Ng, End-to-end text recognition with convolutional neural networks, in: Proceedings of International Conference on Computer Vision and Pattern Recognition, 2012, pp. 3304–3308.

[20] M. Jaderberg, K. Simonyan, A. Vedaldi, A. Zisserman, Reading text in the wild with convolutional neural networks, International Journal of Computer Vision 116 (1) (2016) 1–20.

[21] J. Redmon, S. Divvala, R. Girshick, A. Farhadi, You only look once: Unified, real-time object detection, in: Proceedings of International Conference on Computer Vision and Pattern Recognition, 2016.

[22] W. Liu, D. Anguelov, D. Erhan, C. Szegedy, S. Reed, Ssd: Single shot multibox detector, in: Proceedings of European Conference on Computer Vision, 2016.

[23] Z. Zhong, L. Jin, S. Huang, Deeptext: A new approach for text proposal generation and text detection in natural images, in: IEEE International Conference on Acoustics, 2017.

[24] A. Gupta, A. Vedaldi, A. Zisserman, Synthetic data for text localisation in natural images, in: Proceedings of International Conference on Computer Vision and Pattern Recognition, 2016.

[25] M. Liao, B. Shi, B. Xiang, Textboxes++: A single-shot oriented scene text detector, IEEE Transactions on Image Processing 27 (8) (2018) 3676–3690..

[26] J. Ma, W. Shao, Y. Hao, W. Li, W. Hong, Y. Zheng, X. Xue, Arbitrary oriented scene text detection via rotation proposals, IEEE Transactions on Multimedia PP (99) (2017) 1–1.

[27] B. Shi, X. Bai, S. Belongie, Detecting oriented text in natural images by linking segments, in: Proceedings of International Conference on Computer Vision and Pattern Recognition, 2017.

[28] P. Lyu, Y. Cong, W.Wu, S. Yan, B. Xiang, Multi-oriented scene text detection via corner localization and region segmentation, in: Proceedings of International Conference on Computer Vision and Pattern Recognition, 2018.

[29] J.Wang, K. Chen, S. Yang, C. C. Loy, D. Lin, Region proposal by guided anchoring, in: Proceedings of International Conference on Computer Vision and Pattern Recognition, 2019.

[30] T. Y. Lin, P. Doll, R. Girshick, K. He, B. Hariharan, S. Belongie, Feature pyramid networks for object detection, in: Proceedings of International Conference on Computer Vision and Pattern Recognition, 2017.

[31] J. Dai, H. Qi, Y. Xiong, Y. Li, G. Zhang, H. Hu, Y. Wei, Deformable convolutional networks, in: Proceedings of International Conference on Computer Vision, 2017.

[32] Z. Zhong, L. Sun, Q. Huo. An Anchor-Free Region Proposal Network for Faster R-CNN based Text Detection Approaches[J]. Document Analysis & Recognition, 2018.

[33] C. Yao, X. Bai, N. Sang, X. Zhou, S. Zhou, and Z. Cao, "Scene text detection via holistic, multi-channel prediction," arXiv preprint arXiv:1606.09002, 2016.

[34] Pan He, Weilin Huang, Tong He, Qile Zhu, Yu Qiao, and Xiaolin Li. Single shot text detector with regional attention. In ICCV, 2017.

[35] Tian Z, Huang W, He T, et al. Detecting Text in Natural Image with Connectionist Text Proposal Network[J]. 2016.

[36] P. Lyu, M. Liao, C. Yao, W. Wu, and X. Bai. Mask textspotter: An end-to-end trainable neural network for spotting text with arbitrary shapes. arXiv preprint arXiv:1807.02242, 2018. 1, 2, 6, 7

[37] W. He, Y. Zhang, F. Yin, and C. Liu. Deep direct regression for multi-oriented scene text detection. In CVPR, pages 745–753, 2017. 1, 6

[38] X. Zhou, C. Yao, H. Wen, Y. Wang, S. Zhou, W. He, and J. Liang. East: an efficient and accurate scene text detector. In CVPR, pages 2642–2651, 2017. 1, 6

[39] Long, J. Ruan, W. Zhang, X. He, W. Wu, and C. Yao. Textsnake: A flexible representation for detecting text of arbitrary shapes. arXiv preprint arXiv:1807.01544, 2018. 1, 2, 6

[40] M. Liao, B. Shi, X. Bai, X. Wang, and W. Liu. Textboxes: A fast text detector with a single deep neural network. In AAAI, pages 4161–4167, 2017. 2

[41] P. He, W. Huang, T. He, Q. Zhu, Y. Qiao, and X. Li. Single shot text detector with regional attention. In ICCV, volume 6, 2017. 1, 2, 6

[42] Y. Jiang, X. Zhu, X. Wang, S. Yang, W. Li, H. Wang, P. Fu, and Z. Luo. R2cnn: rotational region cnn for orientation robust scene text detection. arXiv preprint arXiv:1706.09579, 2017. 1, 6